\pdfoutput=1

\documentclass[runningheads]{llncs}

\usepackage[utf8]{inputenc}
\usepackage{graphicx,verbatim}
\usepackage{soul}
\usepackage{xcolor}
\usepackage{hyperref}
\usepackage{booktabs}

\sethlcolor{yellow}

\usepackage{amsmath,amssymb}

\usepackage{times}
\usepackage{latexsym}

\usepackage{lineno}

\usepackage{graphicx} 

\usepackage[T1]{fontenc}

\usepackage[utf8]{inputenc}

\usepackage{microtype}

\usepackage{inconsolata}

\usepackage{soul}
\usepackage{xcolor}
\sethlcolor{yellow}

 \usepackage{natbib}

\title{User Perception of Attention Visualizations: Effects on Interpretability Across Evidence-Based Medical Documents}

\author{
Andrés Carvallo\inst{1} \and
Denis Parra\inst{2} \and
Peter Brusilovsky\inst{3} \and
Hernan Valdivieso\inst{2} \and
Gabriel Rada\inst{2} \and
Ivania Donoso\inst{4} \and
Vladimir Araujo\inst{2}
}

\institute{
CENIA – Centro Nacional de Inteligencia Artificial, Chile\\
\email{afcarvallo@uc.cl}
\and
Pontificia Universidad Católica de Chile, Chile\\
\email{\{dparras,hfvaldivieso,vgaraujo\}@uc.cl}, \email{gabriel@rada.cl}
\and
University of Pittsburgh, United States\\
\email{peterb@pitt.edu}
\and
KU Leuven, Belgium\\
\email{indonoso@uc.cl}
}

\titlerunning{Attention Visualizations in Biomedical NLP}
\authorrunning{Carvallo et al.}

\begin{document}

\maketitle

\begin{abstract}
The attention mechanism is a core component of the Transformer architecture. Beyond improving performance, attention has been proposed as a mechanism for explainability via attention weights, which are associated with input features (e.g., tokens in a document). In this context, larger attention weights may imply more relevant features for the model’s prediction. In evidence-based medicine, such explanations could support physicians' understanding and interaction with AI systems used to categorize biomedical literature. However, there is still no consensus on whether attention weights provide helpful explanations.
Moreover, little research has explored how visualizing attention affects its usefulness as an explanation aid. To bridge this gap, we conducted a user study to evaluate whether attention-based explanations support users in biomedical document classification and whether there is a preferred way to visualize them. The study involved medical experts from various disciplines who classified articles based on study design (e.g., systematic reviews, broad synthesis, randomized and non-randomized trials).
Our findings show that the Transformer model (XLNet) classified documents accurately; however, the attention weights were not perceived as particularly helpful for explaining the predictions. However, this perception varied significantly depending on how attention was visualized. Contrary to Munzner’s principle of visual effectiveness, which favors precise encodings like bar length, users preferred more intuitive formats, such as text brightness or background color. While our results do not confirm the overall utility of attention weights for explanation, they suggest that their perceived helpfulness is influenced by how they are visually presented.
\end{abstract}

\section{Introduction}

Transformers \citep{vaswani2017attention} have achieved state-of-the-art results across a wide range of tasks, including Natural Language Processing (NLP) \citep{canchila2024natural}, Computer Vision \citep{khan2022transformers}, and Information Retrieval \citep{wang2024utilizing}. 

Despite their success, Transformers are often criticized for their lack of interpretability. Their complex architectures, involving millions of parameters, make it difficult to understand the underlying reasoning behind their predictions.

The attention mechanism \citep{bahdanau2014neural}, introduced initially to enhance performance in sequence-to-sequence models, has become a central component of Transformers. Self-attention enables models to capture contextual relationships by assigning weights to input elements, such as tokens in a document. These attention weights have been proposed as a potential form of explanation for model predictions \citep{parra2019analyzing}.

However, there is an ongoing debate about whether attention weights truly provide meaningful or trustworthy explanations \citep{jain2019attention}. Moreover, limited research has investigated whether certain ways of visualizing attention in text are perceived by users as more helpful.

This issue is particularly relevant in evidence-based medicine (EBM), where clinicians must quickly assess large volumes of literature to support medical decisions \citep{elliott2014living}. In such settings, AI systems must not only be accurate but also provide intuitive, trustworthy explanations that help users work more efficiently.

To address this gap, we:
\begin{enumerate}
    \item Developed a system that classifies biomedical research articles and generates visual explanations based on attention weights from a Transformer model.
    \item Conducted a user study to evaluate whether attention weights and the model's predicted probability are perceived as helpful explanations in biomedical document classification.
    \item Compared different ways of visualizing attention in text and assessed whether their perceived usefulness varies depending on the type of document being reviewed.
\end{enumerate}

\section{Related Work}

\subsection{Attention as Explanation}

The use of attention weights has been proposed as a means to interpret Transformer-based models \citep{parra2019analyzing}; however, their ability to explain predictions remains debated. While some argue that attention weights do not reflect model reasoning \citep{jain2019attention}, others support their utility under certain conditions \citep{wiegreffe2019attention}. Recent studies even question the role of attention altogether, pointing instead to feed-forward layers \citep{geva2022transformer}. Despite this, attention visualizations remain relevant in biomedical NLP, where encoder-based models fine-tuned on domain-specific corpora can yield interpretable patterns \citep{roccabruna2024will}. Prior work in biomedical text classification has used Transformer-based models \citep{carvallo2020neural,carvallo2020automatic,carvallo2019comparing} and investigated their robustness \citep{araujo2020adversarial,aspillaga2020stress,araujo2021stress}. In this work, we build upon these foundations to explore how attention visualization affects perceived usefulness in medical document classification.

\subsection{Interfaces for Attention Visualization}

Tools like \texttt{BertViz} \citep{vig2019bertviz} and \texttt{AttentionViz} \citep{yeh2023attentionviz} enable users to inspect attention weights across layers and heads. However, they rarely evaluate how attention shown \emph{within the text} affects human perception. Our study complements this line of work by conducting a user evaluation focused on perceived usefulness of attention visualizations in the text, across different types of biomedical evidence. We also build on previous applications in evidence-based medicine \citep{carvallo2023comparative} and adversarial evaluation in biomedical NLP tasks \citep{araujo2020adversarial2} to assess how attention-based explanations perform in realistic, high-stakes settings.

\subsection{User Studies on Explainability}

User-centered XAI research indicates that the usefulness of explanations depends on user expertise, control, and context \citep{cai2019effects, eiband2019people}. Recent studies highlight the importance of aligning explanation design with domain-specific needs and user profiles, especially in healthcare \citep{e2024evaluating, kim2023should}. In this work, we contribute to this line of research by evaluating how medical experts perceive different visual explanations of attention and whether such visualizations support their task of classifying biomedical evidence.

\section{The Explainable Interface}

We developed an interface within the Epistemonikos \footnote{\url{https://www.epistemonikos.org/}} platform to enable interaction with a Transformer-based model that highlights word-level attention scores. Epistemonikos, a non-profit focused on evidence-based medicine, is widely used by physicians. To preserve its original workflow, we deployed our interface as a Chrome extension\footnote{\url{https://chromewebstore.google.com/category/extensions}} that overlays visual explanations without modifying the underlying system.

\begin{figure}[!h]
\includegraphics[width=0.95\columnwidth]{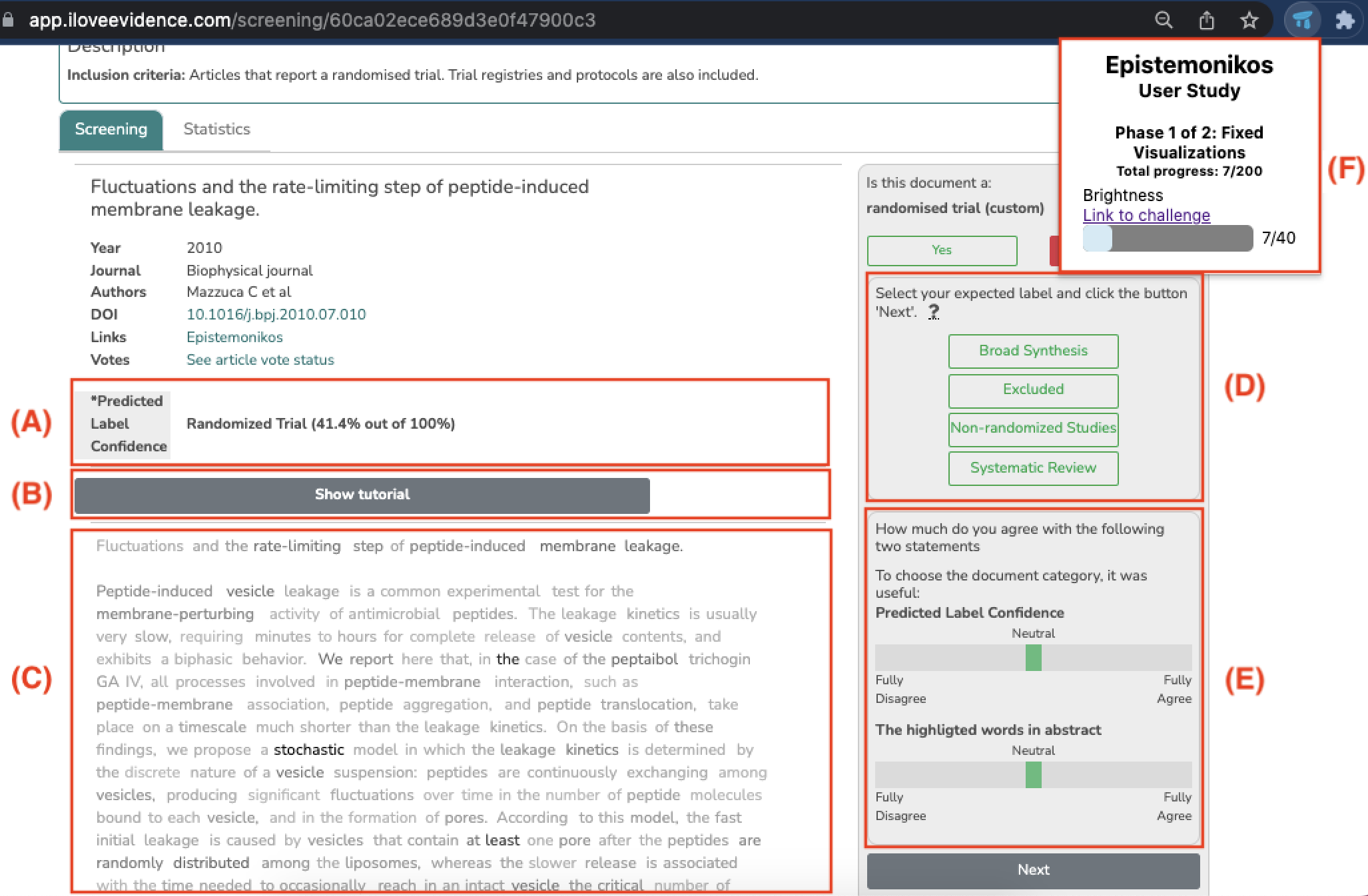}
  \caption{ 
  Screenshot of the Epistemonikos user study interface used to evaluate attention-based explanations in biomedical document classification. (A) Model-predicted label and confidence score. (B) Interactive tutorial toggle. (C) Biomedical abstract with attention-based word highlighting (e.g., via brightness). (D) User Label Selection Options for Document Classification. (E) Likert-scale feedback on the perceived usefulness of the explanation components. (F) Study progress and current visualization condition. }~\label{interface}
\end{figure}

Figure~\ref{interface} shows the proposed interface with six key components: \textbf{(A)} the model’s predicted study type, \textbf{(B)} a help/tutorial button, \textbf{(C)} the abstract with word-level attention highlights, \textbf{(D)} user label selection, \textbf{(E)} feedback on the predicted label and highlighted words, and \textbf{(F)} a progress bar. The interface enables the comparison of different attention visualizations, designed in accordance with the \textit{effectiveness principle} in information visualization, which prioritizes perceptually accurate encodings of key information \citep{midway2020principles}.

\begin{figure}
\includegraphics[width=0.9\columnwidth]{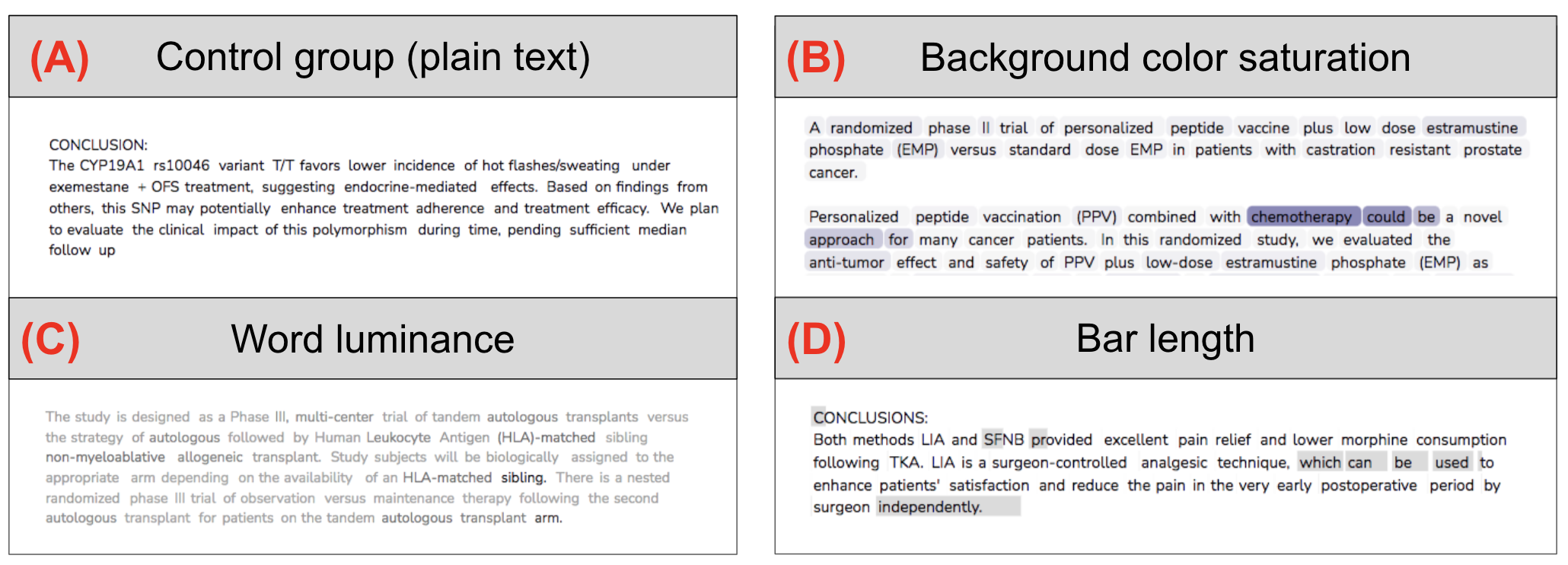}
  \caption{Examples of the four ways of visualizing attention on text evaluated in the study. (A) Plain text (control condition with no attention cues), (B) background color saturation, (C) word luminance, and (D) bar length below each word.}~\label{visual_encodings}
\end{figure}

We tested three different ways of visualizing attention in text, along with a control condition without visualization, as shown in Figure~\ref{visual_encodings}. In the control group (A), the abstract is shown as plain text with no visual cues. In the background color condition (B), each word's background is shaded based on its attention weight—the darker the background, the more important the word. In the word luminance condition (C), text brightness varies with attention, making relevant words appear darker. In the bar length condition (D), a horizontal bar is displayed beneath each word, proportional to its importance. In addition to attention visualization, we also tested whether displaying the model’s predicted probability (or certainty) helped users make classification decisions. We analyzed the results across different types of evidence-based medical documents—such as randomized trials, systematic reviews, and broad syntheses—to evaluate how both attention visualization and model confidence influence users’ perceived usefulness and decision-making.

\section{Language Models}

We evaluated three attention-based encoder models: BERT \citep{devlin2018bert}, BioBERT \citep{lee2020biobert}, and XLNet \citep{yang2019xlnet}. These models were fine-tuned for a multi-class classification task over medical literature within the context of evidence-based medicine (EBM). The classification was performed by passing the special [CLS] token through a fully-connected layer followed by a softmax activation, producing a probability distribution over five evidence types: Broad Synthesis (BS), Excluded (EXC), Randomized Controlled Trial (PS-RCT), Non-Randomized Controlled Trial (PS-NRCT), and Systematic Review (SR). SR and PS-RCT represent the highest levels in the hierarchy of medical evidence \citep{gopalakrishnan2013systematic}. 

We chose encoder-based models over large autoregressive language models (LLMs) due to their efficiency in inference, stability during fine-tuning, and direct interpretability through attention weights. Encoders are particularly suitable for classification tasks over fixed-length inputs, and their self-attention mechanisms produce structured outputs that are easier to align with human-interpretable features \citep{roccabruna2024will}.

\begin{table*}[!ht]
\centering
\resizebox{0.9\textwidth}{!}{%
\begin{tabular*}{\textwidth}{@{\extracolsep{\fill}}lccc|ccc|ccc}
\toprule
\textbf{Type} & \multicolumn{3}{c|}{\textbf{BERT}} & \multicolumn{3}{c|}{\textbf{XLNet}} & \multicolumn{3}{c}{\textbf{BioBERT}} \\
& \textbf{Prec.} & \textbf{Rec.} & \textbf{F1} & \textbf{Prec.} & \textbf{Rec.} & \textbf{F1} & \textbf{Prec.} & \textbf{Rec.} & \textbf{F1} \\
\midrule
BS       & 0.53 & 0.37 & 0.44 & \textbf{0.84} & \textbf{0.77} & \textbf{0.81} & 0.56 & 0.69 & 0.62 \\
EXC      & 0.86 & 0.83 & 0.84 & \textbf{0.97} & \textbf{0.96} & \textbf{0.97} & 0.90 & 0.62 & 0.73 \\
PS-RCT   & 0.63 & 0.84 & 0.72 & \textbf{0.83} & \textbf{0.89} & \textbf{0.86} & 0.64 & 0.80 & 0.71 \\
PS-NRCT  & 0.91 & 0.93 & 0.92 & \textbf{0.99} & \textbf{0.99} & \textbf{0.99} & 0.82 & 0.96 & 0.88 \\
SR       & 0.90 & 0.93 & 0.91 & \textbf{0.94} & \textbf{0.97} & \textbf{0.96} & \textbf{0.94} & 0.92 & 0.93 \\
\midrule
\textbf{Avg} & 0.88 & 0.88 & 0.88 & \textbf{0.97*} & \textbf{0.97*} & \textbf{0.97*} & 0.85 & 0.84 & 0.84 \\
\bottomrule
\end{tabular*}%
}
\caption{Results obtained for document classification across five biomedical evidence types. Best-performing values are in bold. The * symbol denotes statistical significance based on the Friedman ad-hoc test.}
\label{models_finetuned_espitemonikos}
\end{table*}

Table \ref{models_finetuned_espitemonikos} shows the performance of the three models on a large-scale EBM dataset, composed of 399,737 documents for training and 18,854 for testing, sourced from Epistemonikos. XLNet consistently outperformed BERT and BioBERT across all evidence categories, with statistically significant gains in precision, recall, and F1-score. Given its superior performance, we selected XLNet to provide attention weights for the explainable interface. These weights were extracted from the final encoder layer and averaged across attention heads to produce a word-level importance score used in visualizations.

\section{Study Design}

We designed a user study to investigate how different explanation components affect user perception during the classification of biomedical documents. Specifically, we examined three factors: \textit{(1) whether attention-based explanations are perceived as helpful, (2) whether certain ways of visualizing attention in text are preferred or more effective, and (3) whether the model’s predicted probability (or certainty) supports decision-making}. 

These aspects were evaluated across multiple types of evidence-based medical articles, including systematic reviews, randomized trials, and non-randomized studies.

The study consisted of two phases:

\textbf{Phase one} was a controlled experiment in which participants used our explainable interface to classify articles. After each classification, they rated the usefulness of the model’s predicted probability and the attention-highlighted words using 5-point Likert scales.

\textbf{Phase two} allowed participants to choose their preferred method of visualizing attention—or disable it entirely—and continue classifying documents under their selected setting. This phase captured user preferences in a more flexible interaction scenario.

The study involved \textbf{five medical experts from diverse specialties}, each of whom labeled 200 biomedical articles, resulting in 1,000 annotated records. The task reflected a realistic evidence-based medicine (EBM) setting, where clinicians categorize literature by study design and quality.

Attention was visualized in multiple formats, as described in Section~\ref{visual_encodings}. After each article, participants answered:
(1) \textit{On a scale from 1 to 5, how helpful was the model’s predicted probability in classifying this article?}  
(2) \textit{On a scale from 1 to 5, how helpful were the highlighted words in the abstract?}

\section{Results}

We analyzed the relationship between users' perceived helpfulness of model explanations both the predicted probability and the attention-based highlighted words—across different article types and visual encodings. This was done using a two-way ANOVA to explore interaction effects.

\begin{figure}
  \includegraphics[width=0.8\columnwidth]{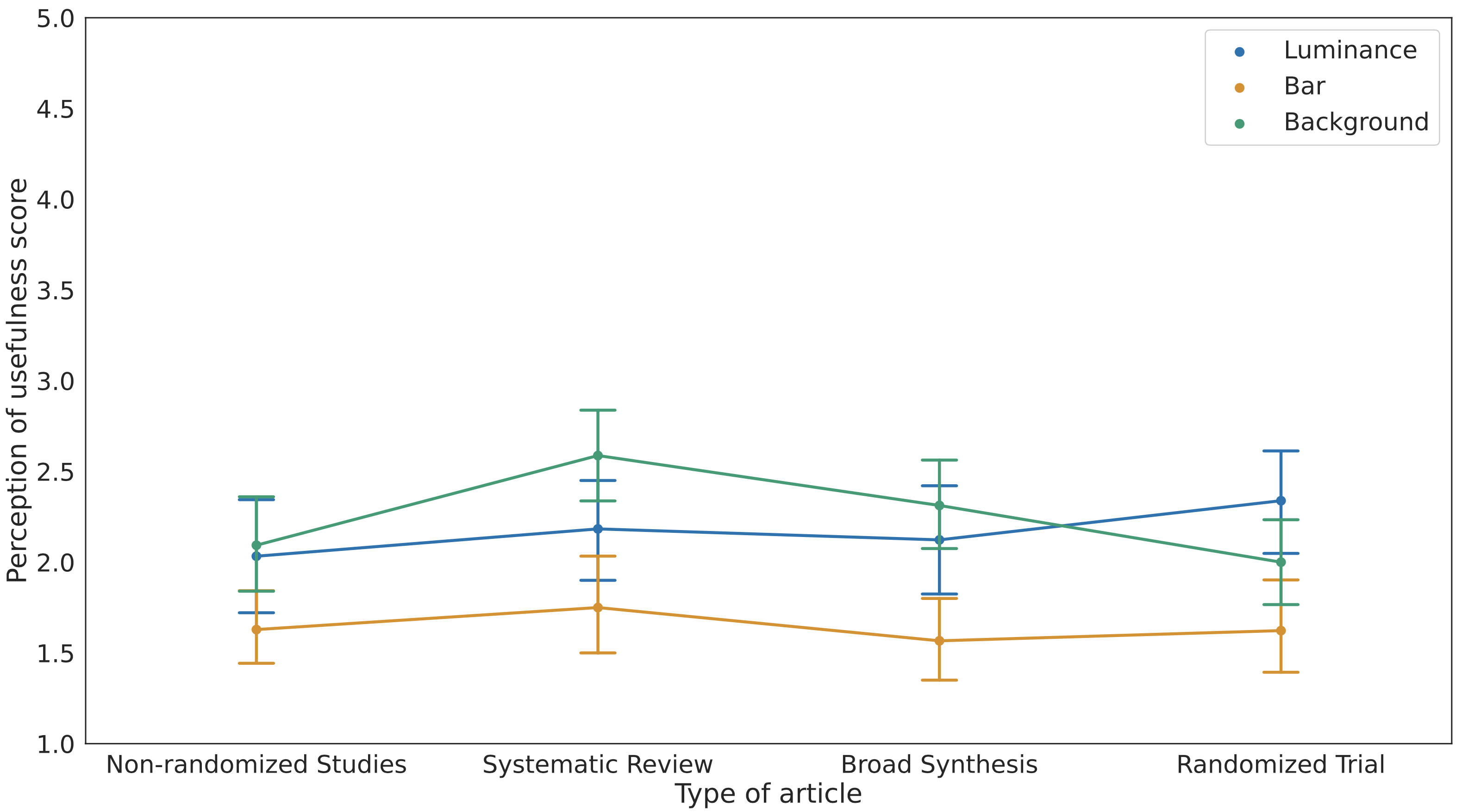}
  \caption{
  Perceived usefulness of highlighted words by visualization type and article category. A two-way ANOVA revealed that usefulness ratings for background color and luminance varied according to the type of document, with higher ratings for Systematic Reviews and Broad Syntheses. In contrast, the bar encoding was consistently perceived as less useful across all types.} ~\label{anova_words}
\end{figure}

Figure~\ref{anova_words} shows the interaction effect between article type and visual encoding on the perceived helpfulness of \textit{highlighted words} (i.e., attention-based explanations). Overall, users rated the usefulness of highlighted words relatively low, with scores peaking around 3.0 on a 5-point scale.

We found a significant interaction effect indicating that the perceived helpfulness of explanations depended on both the type of article and the visual encoding used. For Systematic Reviews (SR), Background encoding was rated as more helpful (M=2.58, SD=1.15) than Bar encoding (M=1.75, SD=1.09). In Broad Synthesis (BS) articles, both Luminance (M=2.12, SD=1.15) and Background (M=2.32, SD=1.14) encodings were perceived as more helpful than Bar (M=1.57, SD=0.89). For Randomized Controlled Trials (PS-RCT), Luminance encoding (M=2.34, SD=1.14) was also rated significantly higher than Bar (M=1.62, SD=1.04). Similarly, in Non-Randomized Controlled Trials (PS-NRCT), Background (M=2.09, SD=1.14) and Luminance (M=2.03, SD=1.22) encodings were rated more helpful than Bar (M=1.63, SD=0.82). 

These results suggest that users generally found Bar Length—despite being the most perceptually accurate channel—less helpful than more intuitive encodings like Background or Luminance. This contradicts the expected effectiveness hierarchy in visualization literature.

\begin{figure}
  \includegraphics[width=0.8\columnwidth]{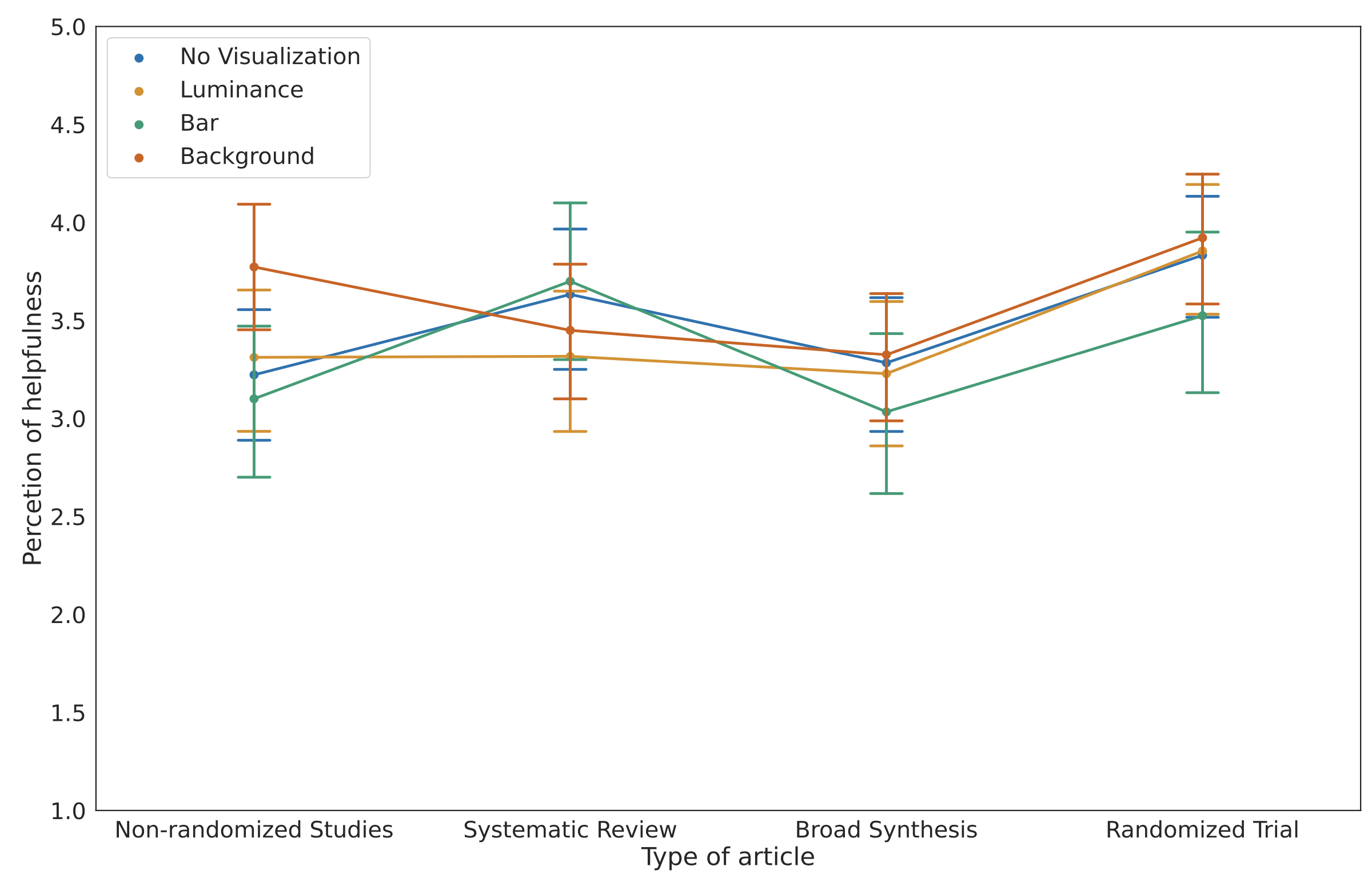}
  \caption{Two-way ANOVA analyzing the interaction between article type and attention visualization on the perceived usefulness of the model's predicted probability. Results indicate that showing the model's probability is consistently perceived as helpful, regardless of the type of document being classified or how attention is visualized.}
  \label{anova_proba}
\end{figure}

Figure~\ref{anova_proba} presents the results of the second ANOVA, analyzing the perceived usefulness of the \textit{model’s predicted probability}. In contrast to attention-based explanations, the predicted probability was consistently rated as highly useful across all article types, with mean scores above 4.0. Importantly, we found no statistically significant differences between visual encodings (including the no-visualization condition) in terms of the perceived usefulness of predicted probabilities. This suggests that while users rely on probability information, their perception of its utility is not influenced by how other explanations are visually presented. Overall, these findings indicate that attention-based explanations are more sensitive to their visual representation and the nature of the underlying content. In contrast, predicted probabilities are perceived as consistently helpful, regardless of the visual context.

\begin{table}[ht]
\centering
\begin{tabular}{|l|c|c|c|c|c|c|}
\hline
\textbf{Visual encoding} & \textbf{Mental} & \textbf{Physical} & \textbf{Temporal} & \textbf{Performance} & \textbf{Effort} & \textbf{Frustration} \\
\hline
No visualization & 46.1 (25.15) & 25.3 (11.08) & 44.2 (23.54) & \textbf{61.6} (16.87) & 49.70 (27.32) & \textbf{27.8} (16.44) \\
Background color & \textbf{37.2} (26.81) & \textbf{24.4} (19.74) & \textbf{36.5} (26.12) & 55.3 (28.59) & \textbf{42.4} (27.73) & 30.3 (25.05) \\
Word luminance & 49.1 (30.54) & 35.1 (26.13) & 49.4 (27.51) & 50.6 (25.57) & 54.3 (30.71) & 43.3 (27.98) \\
Bar length & 48.5 (24.28) & 35.4 (22.82) & 52.5 (23.58) & 59.1 (15.58) & 56.5 (24.95) & 49.4 (21.66) \\
\hline
\end{tabular}
\caption{Mean (standard deviation) for NASA-TLX subscales across visual encoding conditions. Bolded values represent the best (lowest workload or highest performance) scores.}
\label{tab:nasa_tlx}
\end{table}


Table~\ref{tab:nasa_tlx} shows the average NASA-TLX scores across different ways of visualizing attention in text. The background color condition was associated with the lowest mental demand (37.2), physical demand (24.4), temporal demand (36.5), and effort (42.4), indicating a lower overall cognitive load compared to other visualization methods. Although the no-visualization condition yielded the highest perceived performance (61.6) and the lowest frustration (27.8), it also showed higher mental demand (46.1) and temporal demand (44.2) than the background color. In contrast, bar length and word luminance produced higher scores across all workload dimensions, with bar length showing the highest frustration (49.4) and effort (56.5). 

These results suggest that background color provides a favorable balance between interpretability and cognitive effort. The high performance and low frustration observed in the no-visualization condition may reflect users’ familiarity with the traditional Epistemonikos interface, while more complex or unfamiliar formats appear to increase cognitive load.


\section{Conclusions}

This study evaluated whether attention-based explanations and predicted probabilities support medical experts in classifying biomedical literature. Attention weights were generally not perceived as helpful, and their usefulness varied depending on how they were visualized and on the type of document being classified. Simpler visualizations, such as background color, were preferred over more precise but cognitively demanding ones like bar length. In contrast, predicted probability was consistently perceived as helpful across all visualization settings and document types.

NASA-TLX results reinforced these findings: background color was associated with lower cognitive load, while bar length and luminance increased effort and frustration. The high performance and low frustration in the no-visualization condition likely reflect user familiarity with the standard interface for evidence-based medicine annotation.

\textbf{Limitations} include the focus on a single domain, a small sample size, and the use of one explanation mechanism.

\textbf{Future work} will explore interactive explanations and extend the study to broader user groups and medical tasks.

\section*{Disclosure of Interests}
The authors declare that they have no competing interests.

\section*{Acknowledgments}
This work was supported by ANID Basal Fund, National Center for Artificial Intelligence CENIA FB210017, Millennium Science Initiative code ICN2021\_004 (iHealth), Postdoctoral FONDECYT grant 3240001, and FONDECYT regular grant 1231724.

\bibliography{anthology,custom}
\bibliographystyle{acl_natbib}

\end{document}